\documentclass{article}
\usepackage{arxiv}

\usepackage[utf8]{inputenc} 
\usepackage[T1]{fontenc}    
\usepackage{hyperref}       
\usepackage{url}            
\usepackage{booktabs}       
\usepackage{amsfonts}       
\usepackage{nicefrac}       
\usepackage{microtype}      
\usepackage{lipsum}		
\usepackage{graphicx}
\usepackage{doi}

\usepackage{subfigure}

\usepackage{algorithmic}
\usepackage{amsthm}
\usepackage{thmtools,thm-restate}
\usepackage{amsmath}
\usepackage{amssymb}
\usepackage{algorithm}
\usepackage{enumitem}
\usepackage{adjustbox}
\usepackage{amsthm}
\usepackage[switch]{lineno}
\newtheorem{thm}{Theorem}
\newtheorem{definition}{Definition}

\newtheorem{property}{Property}

\newcommand{\loss}{\mathcal L \space }
\newcommand{\wubound}{\hat{\Delta}_S}

\newcommand{\empbound}{\hat{\Delta}_S^*}
\newcommand{\wu}{\Delta_S}
\newcommand{\var}{\Delta_V}
\newcommand{\wuvar}{\Delta_{S+V}}
\newcommand{\epsi}{\epsilon_i(\mathcal{D})}
\newcommand{\sigmai}{\sigma_i(\mathcal{D})}

\title{An Empirical Study on the Intrinsic Privacy of Stochastic Gradient Descent}

\author{ \href{https://orcid.org/0000-0001-7571-6036}{\includegraphics[scale=0.06]{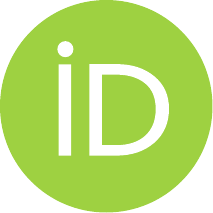}\hspace{1mm}Stephanie L. Hyland}\\
	Microsoft Research Cambridge\\
	\texttt{stephanie.hyland@microsoft.com} \\
	\And
	Shruti Tople \\
	Microsoft Research Cambridge\\
	\texttt{shruti.tople@microsoft.com}
}

\date{}

\renewcommand{\headeright}{Technical Report}
\renewcommand{\undertitle}{Technical Report}
\renewcommand{\shorttitle}{An Empirical Study on the Intrinsic Privacy of SGD}

\begin{document}
\maketitle





\begin{abstract}
{Introducing noise in the training of machine learning systems is a powerful way to protect individual privacy via differential privacy guarantees, but comes at a cost to utility.
This work looks at whether the inherent randomness of stochastic gradient descent (SGD) could contribute to privacy, effectively reducing the amount of \emph{additional} noise required to achieve a given privacy guarantee.
We conduct a large-scale empirical study to examine this question.
Training a grid of over 120,000 models across four datasets (tabular and images) on convex and non-convex objectives, we demonstrate that the random seed has a larger impact on model weights than any individual training example.
We test the distribution over weights induced by the seed, finding that the simple convex case can be modelled with a multivariate Gaussian posterior, while neural networks exhibit multi-modal and non-Gaussian weight distributions.
By casting convex SGD as a Gaussian mechanism, we then estimate an `intrinsic' data-dependent $\epsi$, finding values as low as 6.3, dropping to 1.9 using empirical estimates.
We use a membership inference attack to estimate $\epsilon$ for non-convex SGD and demonstrate that hiding the random seed from the adversary results in a statistically significant reduction in attack performance, corresponding to a reduction in the effective $\epsilon$.
These results provide empirical evidence that SGD exhibits appreciable variability relative to its dataset sensitivity, and this `intrinsic noise' has the potential to be leveraged to improve the utility of privacy-preserving machine learning.}
\end{abstract}

\section{Introduction}
Respecting the privacy of people contributing their data to train machine learning models is important for the safe and responsible use of this technology~\cite{shokri2017membership,tramer2016stealing,fredrikson2015model}.
Private variants of learning algorithms have been proposed to address this need~\cite{chaudhuri2011differentially,wu2017bolt,rajkumar2012differentially,song2013stochastic,geyer2017differentially}.
Unfortunately the utility of private models is typically degraded, limiting their applicability.
This performance loss often results from the need to add noise to provide the strong protections of $\epsilon$-differential-privacy~\cite{dwork2011differential}.

Missing in approaches to date is the observation that learning algorithms \emph{themselves} often feature noise.
Randomness of optimisation is well-known for users of stochastic gradient descent (SGD).
In SGD, randomness arising from model initialisation and minibatch sampling can yield dramatically different results~\cite{Henderson2017DeepRL,Frankle2018TheLT}, which is often perceived as a limitation, even if the stochasticity itself is credited with improving generalisation over batch gradient descent~\cite{Keskar2016OnLT}.
To date, SGD has been framed as a `fixed' query on a dataset, neglecting an important source of \emph{intrinsic} noise.
However, if we could quantify and account for this intrinsic noise, the amount of \emph{additional} noise required to achieve a target level of privacy would be reduced, yielding likely improvements in model performance.

\textbf{Idea: variability versus stability.}
Our motivating scenario is one where a model is trained securely, but the final model parameters are released to the public - for example, a hospital which trains a prediction model on its own patient data and then shares it with other hospitals or a cloud provider.
We therefore focus on how SGD introduces randomness in the \emph{final} weights of a trained model.

\begin{figure}[t]
\centering
    \includegraphics[scale=0.75]{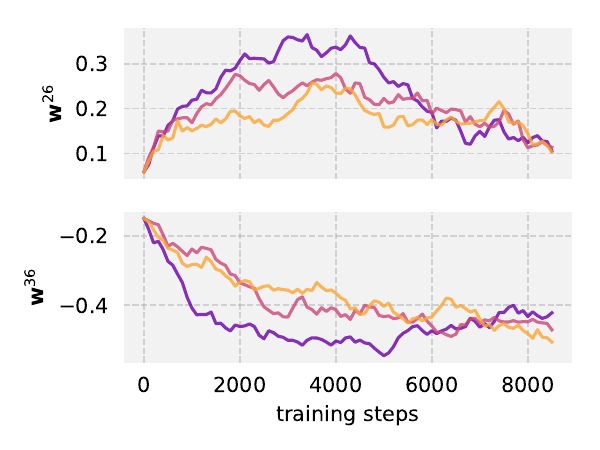}
    \caption{A demonstration of the variability of SGD due to its random seed.
Each curve shows the evolution model weights $\mathbf{w}^{26}$ and $\mathbf{w}^{36}$ throughout training, using runs of SGD on the same data with identical initialisation, with different random seeds and thus data sampling order.
The model is logistic regression on the \texttt{Forest} dataset.}
    \label{fig:intuition}
\end{figure}

This randomness manifests primarily in two ways:
\begin{enumerate}
    \item Initialization of the model parameters
    \item Sampling of the minibatch samples while training
\end{enumerate}
We highlight that training a model on \emph{the same data}, even with the \emph{same initialisation} can output different weights, as shown in Figure~\ref{fig:intuition}.

At the same time, prior work on the uniform stability of SGD~\cite{hardt2015train,Kuzborskij2017DataDependentSO} has highlighted that small variations in the training data provided to SGD produce theoretically bounded changes in the resulting model.
Combining these observations, we investigate whether the \emph{variability} in model parameters induced by the stochasticity of SGD could exceed its \emph{sensitivity} to perturbations in the input data.
Rather than viewing this variability as a pitfall of stochastic optimisation, it could instead be seen as a source of noise that can mask information about participants in the training data, affording some data and model dependent \emph{intrinsic} privacy.

\textbf{Approach: empirical study.}
To explore the role of the seed and data in the output of SGD, we conduct a large-scale \emph{empirical} analysis.
Considering four datasets and two model classes (convex and non-convex objectives), we systematically vary either individual data examples or the random seed given to SGD, considering both fixed and variable model initialisation.
This results in 9,000--25,000 trained models per setting, for a total of over 120,000 models.
Recording model weights throughout training, this suite of results allow us to study both the dataset-sensitivity and variability of SGD.

We are able to track how the largest distance between pairs of model weights evolves throughout training, and compare the impact of the data and the random seed on such distances.
By fixing the data, we can explore the distribution over model weights induced by the random seed, and perform statistical tests against known distributions.
We further estimate empirical $\epsilon$ and $\epsi$ values to characterise the potential for privacy through randomness in SGD, using two approaches:
\begin{enumerate}
    \item For convex objectives, we can draw analogy to the Gaussian mechanism and \emph{directly} estimate an intrinsic data-dependent $\epsi$.
    \item For large models with non-convex objectives, we use membership inference attacks where the adversary does or does not know the random seed, to characterise the effective change in $\epsilon$ arising due to the seed.
\end{enumerate}

\textbf{Contributions.} As our contribution, we provide answers to the following research questions.
\begin{enumerate}
    \item Does the variability due to random sampling exceed the sensitivity due to changing a single input? \\
     (Ans): The seed plays a much larger role in determining model weights than any one training example, with existing theoretical bounds on dataset sensitivity demonstrated to be loose.
    \item  What does the noise distribution of SGD arising due to the random seed look like? \\
    (Ans): Convex objectives yield weight distributions which can modelled with a multivariate Gaussian, but this does not hold for non-convex objectives (e.g. neural networks).
    \item What are the privacy implications, in terms of $\epsilon$, for convex and non-convex  SGD? \\
    (Ans): The intrinsic data-dependent $\epsi$ for convex models (given assumptions) varies by dataset, between 1.9 and 17.7 using the optimistic empirical estimate of dataset sensitivity, which is comparable to $\epsilon$ values seen in practice~\cite{jayaraman2019evaluating}. For non-convex models, concealing the random seed from an adversary statistically significantly reduces the performance of membership inference attack.
\end{enumerate}
Our work shows that although the inherent noise in SGD may not be sufficient by itself for strong privacy guarantees but it has the potential to be leveraged for improving privacy, and should be considered in the design of differentially private machine learning systems.

\section{Problem \& Background  }
\label{section:background}

\subsection{Stochastic Gradient Descent (SGD)} 
SGD and its derivatives are the most common optimisation methods for training machine learning models~\cite{bottou2012stochastic}.
Given a loss function $\loss(\mathbf{w}, (x, y))$ averaged over a dataset, SGD provides a stochastic approximation of the traditional gradient descent method by estimating the gradient of $\loss$ at random inputs.
At step $t$, on selecting a batch of $B$ random samples $\{(x, y)_b\}_{b=1}^B$, the model parameters are updated as:
\begin{equation*}
    \mathbf{w}_{t+1} = \mathbf{w}_t - \eta \frac{1}{B}\sum_{b=1}^B \nabla_{\mathbf{w}} \loss(\mathbf{w}_t, (x, y)_b)
\end{equation*}
where $\eta$ is the (constant) step-size or learning rate, and $w_t$ are the weights of the model at $t$. `Pure' stochastic gradient descent sets $B=1$, however in practice $B>1$ is used to reduce the variance of the gradient estimate and improve the performance of the algorithm.

\subsection{Differential Privacy}
Differential privacy hides the participation of an individual sample in the dataset~\cite{dwork2011differential}.
Informally, it ensures that the presence or absence of a single data point in the input dataset does not appreciably change the output of a differentially private query on that dataset.
It is formally defined as follows: 
\begin{definition}[ $(\epsilon, \delta)$-Differential Privacy] A mechanism $\mathcal{M}$ with
domain $\mathcal{I}$ and range $\mathcal{O}$ satisfies $(\epsilon, \delta)$-differential privacy if for any two neighbouring
datasets $S, S' \in \mathcal{I}$ that differ only in one input and for a set ${E}\subseteq
\mathcal{O}$, we have: 
		$ \Pr(\mathcal{M}(S)\in {E}) \leq e^\epsilon \Pr(\mathcal{M}(S')\in {E}) + \delta $
	\label{defn:dp}
\end{definition}
 $(\epsilon, \delta)$-differential privacy ensures that for all adjacent datasets $S$ and $S'$, the privacy loss of
 any individual data point is bounded by $\epsilon$ with probability at least $1-\delta$~\cite{Dwork2014TheAF}.
A well-established method to design ($\epsilon$, $\delta$)-differentially private mechanism is to add noise proportional to the $\ell_1$ or $\ell_2$ sensitivity of the algorithm's output to a change in a single input sample.
We consider an algorithm (SGD) whose output is the weights of a trained model, thus focus on the $\ell_2$-sensitivity:
\begin{definition} [$\ell_2$-Sensitivity (From Def 3.8 in ~\cite{Dwork2014TheAF}] Let $ f$
be a function that maps a dataset to a vector in $\mathbb{R}^d$. Let $ {S}$, $ {S'}$ be two
datasets such that they differ in one data point. Then the $\ell_2$-sensitivity
of a function $f$ is defined as: 
		  $\Delta_2 (f) =\max_{{S}, {S'}}\| f (S) -  f (S')\|_2$
	 \label{defn:sensitivity}
 \end{definition}

\subsection{Sensitivity of SGD}
Building on the stability results of \cite{hardt2015train}, \cite{wu2017bolt} gave a bound for the \emph{sensitivity} of SGD due to the change in a single training example for a convex, $L$-Lipschitz and $\beta$-smooth loss $\loss$, shown in Equation~\ref{eqn:wubound}.
Let A denote the SGD algorithm using $r$ as the random seed with a batch size of $B$.
The upper bound for sensitivity for $k$ epochs of SGD with learning rate $\eta$ is given by
\begin{equation}
\wubound = \max_r \| A(r; S) - A (r; S')  \| \le  2kL\eta / B
\label{eqn:wubound}
\end{equation}
Assuming the random seed to be fixed, $\wubound$ gives the maximum difference in the model parameters due to the presence or absence of a single input sample.
This allows for a differentially-private version of SGD using output perturbation, however it does not account for intrinsic randomness.

When the \emph{random seed} differs between two runs of SGD, the upper bound for the change in model weights increases by a factor of $T$ (the number of training steps), as \emph{almost every} batch will contain differing samples.

\subsection{Threat Model and Setting}
We focus on the problem of model release, assuming a model can be trained securely `in-house' on sensitive data and then released for public access.
For example, a hospital which wishes to share a model with other hospitals, potentially through a cloud provider.
The training procedure (SGD) is therefore the query, and the full model parameters are the output.
The cloud provider or anyone with access to the model parameters is considered as an adversary.
The adversary has access to the fully trained model, including its architecture, and we assume details of the training procedure are public (e.g. batch size, number of epochs, learning rate), but \emph{not} the random seed used to initialise the model and sample inputs from the dataset.
Thus, we consider a powerful adversary with white-box access to the model.

We assume that each user contributes one sample, and that leaking this membership in the dataset poses a privacy risk to them.
Prior research has shown several attacks such as membership inference and model inversion that can compromise the privacy of the training dataset in this setting~\cite{shokri2017membership,fredrikson2015model}.
Output perturbation, i.e. adding noise to the final model weights can provide privacy guarantees as shown by \cite{wu2017bolt}.
In our setting, the adversary views one of a set of possible models arising from potential random seeds.
This introduces uncertainty over model weights and (potential) plausible deniability for participation in the dataset -- before the use of additional noise.
We focus on seed-dependent randomness and do not consider variability introduced by hiding other hyper-parameters such as batch size and learning rate.

\section{Related Work}
\label{section:related}
\subsection{Differentially-private machine learning}
Ongoing work develops differentially-private variants of training algorithms, including objective perturbation~\cite{chaudhuri2011differentially}, gradient perturbation~\cite{abadi2016deep}, and teacher-student frameworks~\cite{papernot2016semi,papernot2018scalable}.
While these methods typically provide privacy throughout training, we focus on the \emph{final} weights of a model trained securely.
This output perturbation setting is addressed most closely by Wu et al.~\cite{wu2017bolt}.
They treat SGD as a `black box' and inject noise on the final model weights, however they do not consider the intrinsic noise in SGD.
The random sampling of SGD is implicitly used for privacy amplification~\cite{balle2018privacy} in DP-SGD~\cite{abadi2016deep}.
We do not consider randomness due to subsampling \emph{alone}, rather the stochasticity induced by the \emph{order} of sampling the entire dataset, and so subsampling is not directly applicable.

\subsection{Estimating Empirical Epsilon}
Empirical studies have recently been conducted on the privacy of models. Jagielski et al.  use data poisoning to estimate empirical lower bounds for existing approaches~\cite{jagielski2020auditing}. This poisoning attack yields `worst-case' privacy under potentially arbitrary dataset perturbations.  Nasr et al. demonstrate that the theoretical upper bounds are often loose in realistic adversarial setting such as a black-box API based adversary~\cite{nasr2021adversary}. However, they also use adversarial examples to compute tight lower bounds estimates. Our approach uses a grid of neighbouring datasets generated by dropping benign samples from a fixed dataset to understand the effect of the random seed during standard training. Our goal is specifically to capture SGD behaviour under non-adversarial conditions.
Our empirical approach could also be combined with other techniques for calculating data-dependent sensitivity such as smooth sensitivity~\cite{nissim2007smooth}.

\subsection{Stability of SGD}
Our observation that SGD is not greatly influenced by individual training examples is motivated by results on its algorithmic stability, a closely related concept.

Bousquet and Elisseeff~\cite{Bousquet2002StabilityAG} first defined the notion of algorithmic stability for a learning algorithm such as SGD and related it to generalisation error.
Hardt, Recht, and Singer~\cite{hardt2015train} expanded these results, relating them to modern deep learning techniques and hyper-parameters of SGD; Kuzborskij and Lampert~\cite{Kuzborskij2017DataDependentSO} extend the analysis to data-dependent stability and generalisation bounds.
In these cases, and in much work studying SGD, the focus is on understanding the generalisation properties of the algorithm.
While the links between stability and generalisation have already been established~\cite{Bousquet2002StabilityAG,hardt2015train}, and between privacy and generalisation~\cite{nissim2007smooth}, we believe ours is the first work to concretely characterize the noise and quantify the potential for intrinsic privacy in SGD.

\subsection{SGD as a randomised mechanism}
Examining the randomness of SGD lends itself to viewing SGD as a sampling procedure over model weights.
Prior work has often modelled SGD as a stochastic differential equation, assuming Gaussian gradient noise~\cite{Mandt2017StochasticGD,Jastrzebski2018ThreeFI,Zhu2019TheAN,Chaudhari2018StochasticGD}.
By making strong assumptions (i.e. SGD as an Ornstein-Uhlenbeck process~\cite{Mandt2017StochasticGD}), an analytic Gaussian form for the weight posterior can be derived.
This allows SGD to be interpreted as equivalent to performing approximate Bayesian inference for a particular choice of variational distribution.
Recently, \cite{gurbuzbalaban2020heavy} demonstrated rather that SGD exhibits a \emph{heavy-tailed distribution}, theoretically analysing the case of linear regression.
They argue that SGD converges to a symmetric $\alpha$-stable distribution, with $\alpha$ dictating the `heaviness' of the tails.
However, to our knowledge there does not yet exist a general theory of the posterior distribution for SGD.
By exploring it computationally, we aim to bring empirical evidence to this line of research.

\subsection{Initialisation of SGD}
The effect of having a good initialisation for SGD is highlighted by the work on `lottery ticket hypothesis'~\cite{Frankle2018TheLT} and the analysis of ~\cite{Frankle2019LinearMC} showed that initialisation noise can produce very different models.
Neural tangent kernels have also yielded results supporting the importance of a good initialisation of generalisation error~\cite{zhang2020type}.
~\cite{jagielski2020auditing} showed that including variable initialisation empirically improves privacy guarantees, which supports our findings that maintaining uncertainty over the initialisation is critical for the randomness in SGD to provide plausible deniability. This effect is particularly important for non-convex objectives with multiple minima.

\section{Empirical Study Design}
\label{section:empirical_study}
Our primary contribution is a large-scale empirical study, which we analyse in subsequent sections.
We consider four datasets and two model classes (convex and non-convex).
For each dataset, we train a grid of models where we vary the data or (non-exclusively) the random seed.
This allows us to explore sensitivity to dataset perturbations, and the intrinsic randomness in SGD respectively.
Code for these experiments is available from \url{https://github.com/microsoft/intrinsic-private-sgd}.

Our evaluation goals are three-fold: 
\begin{enumerate}
\item Quantify and compare variability of SGD arising from changes in data and/or random seed.
\item Characterise the distribution of noise introduced in the weights due to the randomness in SGD. 
\item Where possible, compute an intrinsic data-dependent $\epsi$ that quantifies the potential privacy that stochasticity in SGD would provide to that dataset.
\end{enumerate}

\subsection{Variation in Data  $\rightarrow$ Sensitivity}
To vary the data, we generate `neighbouring' datasets derived from a given benchmark dataset $\mathcal{D}$ (e.g. two variants of \texttt{MNIST} - we \emph{do not} compare between e.g. a model trained on \texttt{MNIST} and one trained on \texttt{CIFAR}).

A pair of datasets is neighbouring if they differ in one example.
Using an approach similar to \cite{hardt2015train}, we define a set of datasets $\{S_i\}$ where $S_i$ is the original data where its $i^{th}$ entry $\mathbf{x}_i$ has been \emph{replaced} by the first example $\mathbf{x}_0$, which is then dropped.
The derived datasets $S_i$ and $S_j$ then differ in that the former is missing $\mathbf{x}_i$, and latter is missing $\mathbf{x}_j$ (each contains $\mathbf{x}_0$ in the $i$th and $j$th positions respectively).
While creating neighbouring datasets, \emph{replacing} the `missing' entry is necessary to ensure that for a fixed seed (and thus order of dataset traversal), the mini-batches sampled from $S_i$ and $S_j$ differ \emph{only} when they would encounter the $i$th and $j$th elements.

As per definition~\ref{defn:sensitivity}, the sensitivity of SGD is given by the largest $\ell_2$-norm change in model weights obtained from neighbouring datasets.
We can empirically compute (an estimate of) this value for both convex and non-convex models as follows:

First, we compute a `pairwise' sensitivity between models trained with the \emph{same seed} ($r$) on neighbouring datasets ($S_i$, $S_j$):
 \begin{equation}
\Delta_S^{rij} = \| A(r, S_i) - A(r, S_j)\|
	\label{eqn:pairwise_sensitivity}
 \end{equation}
The maximum of $\Delta_S^{rij}$ is the `global' (dataset-specific) sensitivity, which we estimate (using a subset of $i, j, r$), as:
\begin{equation}
    \empbound = {\max}_{i, j, r} \|A(r, S_i) - A(r, S_j)\|
    \label{eqn:empirical_sensitivity}
\end{equation}

Since we run a finite grid of experiments, we estimate $\empbound$ using a subset of possible datasets and random seeds. In section~\ref{section:enough_experiments} we demonstrate how the estimate depends on the size of this subset, which highlights the importance of running enough experiments.

For convex models, we can also compare to the (bound on the) theoretical sensitivity of SGD $\wubound$ estimated by~\cite{wu2017bolt,hardt2015train} (Equation~\ref{eqn:wubound}).

\subsection{Variation in Random Seed $\rightarrow$ Variability}
To vary the random seed, we run each experiment (on a given dataset $S_i$) multiple times with different seeds provided to the random number generator.
We ensure all random number generators are set such that there are no further sources of randomness in our experiments.
The random seed impacts the training procedure by specifying the initialisation of the weights, and the order of traversal of the dataset in each epoch.
For illustration we also consider a variant of this setting where the initialisation of the model is fixed, corresponding to fine-tuning a (poor) public model.

We follow the traditional setting of SGD where a random permutation (determined by the random seed) is applied to the training data at the start of each epoch, and batches of examples are sequentially drawn without replacement.

Each random seed specifies a single draw from the distribution over model weights implied by SGD. Drawing more samples allows us to better capture the properties of this distribution. We examine the impact of varying the number of draws in Section~\ref{section:enough_experiments}.

\subsection{Benchmark datasets.}
We use four data benchmark datasets, each with binary classification tasks:
\begin{itemize}[leftmargin=*]
    \item \texttt{CIFAR2}\cite{Krizhevsky2009LearningML}: The original CIFAR-10 dataset, restricted to classes 0 and 2 (planes and birds) to form a binary classification task (hence \texttt{CIFAR2}).
    \item \texttt{MNIST-binary}\cite{LeCun1998GradientbasedLA}: The original MNIST dataset, restricted to classes 3 and 5.
    \item \texttt{Adult}\cite{Dua:2019}\footnote{https://archive.ics.uci.edu/ml/datasets/Adult}: The task is to predict whether an individual's income exceeds \$50k/year, based on census data from 1994.
    \item \texttt{Forest}\cite{Dua:2019,Dean1998ComparisonON}\footnote{https://archive.ics.uci.edu/ml/datasets/Covertype}: Forest cover type prediction from cartographic information. We restrict to classes 1 and 2.
\end{itemize}

\texttt{Adult} and \texttt{Forest} are tabular datasets. We one-hot encode the categorical-valued features in \texttt{Adult}, dropping the first level.
\texttt{CIFAR2} and \texttt{MNIST-binary} are image datasets, with shapes (32, 32, 3) and (28, 28) respectively.
For \texttt{CIFAR2} we follow the convention of fine-tuning a pre-trained model on a `public' dataset~\cite{papernot2019making} - in this case, we take a ResNet-56 trained on CIFAR-100\footnote{\url{https://github.com/chenyaofo/CIFAR-pretrained-models}} and use its penultimate layer as an `embedding' of \texttt{CIFAR-10}, resulting in examples of size $d = 64$. The Appendix~\ref{app:implementation} contains more details on this process. 
For \texttt{MNIST-binary} we simply project examples using PCA to $d = 50$.
Each dataset is normalised such that $||\mathbf{x}|| \leq 1$ to enable use of Equation~\ref{eqn:wubound}.

\texttt{CIFAR}, \texttt{MNIST} and \texttt{Adult} have pre-specified test sets. 
For \texttt{Forest} we randomly select 15\% of examples to form the test set.
We form a validation set, with which to do early stopping  and model selection, by selecting 10\% of the data remaining after removing the test set in all cases.
The sizes of these datasets are given in Table~\ref{table:datasets_summary}.

\begin{table}[t]
\centering
        \begin{tabular}{l ||r | r|r }
            Dataset                 & $N_\textrm{train}$       & $N_\textrm{test}$ & $d$   \\\hline
            \texttt{CIFAR2}         & 9000     & 2000   & 64* \\
            \texttt{MNIST-binary}   & 10397    & 1902   & $50^{\dagger}$  \\
            \texttt{Adult}          & 29305    & 16281  & 100  \\
            \texttt{Forest}         & 378783   & 74272    & 49  \\
        \end{tabular}
    \caption{Statistics for datasets.
    $d$: dimension of feature vectors. *\texttt{CIFAR2} is originally (32, 32, 3), but is `projected' to $d = 64$ using a ResNet.
    $\dagger$\texttt{MNIST-binary} is originally (28, 28), but projected to $d = 50$ using PCA.}
    \label{table:datasets_summary}
\end{table}

\subsection{Models and Training.}
\label{section:training}
We study convex and non-convex objectives by considering two model classes:
\begin{enumerate}
    \item \textbf{Logistic regression}. With the standard cross-entropy loss, the objective function for logistic regression is convex and Lipschitz with constant $L = \text{sup}_{\mathbf{x}}||\mathbf{x}||$ and smooth with $\beta = \text{sup}_{\mathbf{x}}\|\mathbf{x}\|^2$. We fix datasets to satisfy $||\mathbf{x}|| \leq 1$ and the models have a bias term, so $L = \sqrt{2}$.
    \item \textbf{Neural networks}. Fully-connected neural networks with one hidden layer, using a relu nonlinearity and a sigmoid activation on the output. Here, the Lipschitz constant is unknown and we can't use theoretical sensitivity results.
\end{enumerate}

We train with a fixed learning rate for simplicity and to allow the use of the bound in Equation~\ref{eqn:wubound}~\cite{wu2017bolt}.
We perform early stopping using a validation set to avoid over-fitting, which is undesirable both for privacy and performance reasons.
Otherwise, we did not perform extensive hyper-parameter optimisation.

\begin{table}[t]
\centering
    \begin{tabular}{l | l | l}
        Model & Dataset & Accuracy \\\hline
        \texttt{LogReg} & \texttt{CIFAR2} &  90.8\% \\
        & \texttt{MNIST-binary} &  91.6\% \\
        & \texttt{Adult} & 81.6\% \\
        & \texttt{Forest} & 77.4\% \\ \hline
        \texttt{NN} & \texttt{CIFAR2} & 92.0\% \\
        & \texttt{MNIST-binary} & 98.5\% \\
        & \texttt{Adult} & 82.4\% \\
        & \texttt{Forest} & 77.0\% \\
    \end{tabular}
      \caption{Performance (binary accuracy) of each model on each dataset at its convergence point. Recall that models on \texttt{CIFAR2} benefit from the embedding of the pretrained CIFAR-100 model.}
    \label{table:performance}
\end{table}

All models achieve average accuracies between 77.0\% (\texttt{Forest}) and 98.5\% (\texttt{MNIST-binary}) at their respective convergence points - see Table~\ref{table:performance}.

More details, including model hyper-parameters are provided in Appendix Table~\ref{table:hyperparameters}. The total number of models trained as part of the empirical study is shown in Table~\ref{table:experiments_summary}.

\begin{table}[h]
\centering
    \begin{tabular}{l | l || l | l || l}
        Model & Dataset & $n_r$ & $n_S$ & Total \\\hline
        \texttt{LogReg} & \texttt{CIFAR2} &  103 &  49 & 9900 \\
        & \texttt{MNIST-binary} &  100 & 99 & 19600\\
        & \texttt{Adult} & 126 & 40 & 25700\\
        & \texttt{Forest} & 160 & 36 & 20693 \\ \hline
        \texttt{NN} & \texttt{CIFAR2} & 120 & 40 & 9900 \\
        & \texttt{MNIST-binary} & 138 & 23 & 9800 \\
        & \texttt{Adult} & 101 & 28 & 11250 \\
        & \texttt{Forest} & 101 & 28 & 11247 \\
    \end{tabular}
      \caption{Size and shape of experiment grid for each dataset and model. $n_r$ means the average number of random seeds tested, per dataset variant. $n_S$ means the average number of dataset variants tested, per seed. Thus, the total experiments (per model, per dataset) is approximately $2 n_r n_S$, as it includes running both with and without fixed initialisation.}
    \label{table:experiments_summary}
\end{table}

\section{Random Sampling vs. Dataset Sensitivity}
\label{section:eval_var}

Given the models trained as part of the experimental grid as described in the previous section,  we are able to explore the effect of varying two inputs to the model: the underlying dataset, and the random seed used in SGD.
We consider the distance between the weights of models trained under different scenarios:
\begin{enumerate}
    \item Fixed seed and variable data: $\wu := \|A(r; S) - A(r; S')\|$ (as per Equation~\ref{eqn:pairwise_sensitivity})
    \item Variable seed and fixed data: $\var := \|A(r; S) - A(r'; S)\|$
    \begin{enumerate}
        \item allowing for fixed initialisation ($\var^{\text{fix}}$)
        \item and seed-dependent initialisation ($\var^{\text{vary}}$)
    \end{enumerate}
\end{enumerate}

In the above, $r$ and $r'$ are two random seeds, and $S$ and $S'$ are neighbouring datasets.
$A$ is the SGD algorithm, which outputs a vector of model weights, and we compute the $\ell_2$ norm between pairs of model weight vectors in all cases.
We explore how the variability \emph{due to seed} ($\var^{\text{fix}}, \var^{\text{vary}}$) compares to the data sensitivity $\wu$.

The distance between model weights as a function of training time has previously been characterised for the first case (fixed seed and variable data, $\wu$) in the context of algorithmic stability~\cite{hardt2015train}.
As they highlight, training `faster' (i.e. converging earlier) produces superior generalisation through smaller sensitivity.
This is reflected by the linear dependence on the number of training steps on the theoretical bound $\wubound$.
However, the relationship between training time and the (seed-dependent) \emph{variability}, as well as the empirical sensitivity ($\empbound$) is not known.

\begin{figure*}[h]
\centering
    \includegraphics[scale=0.46]{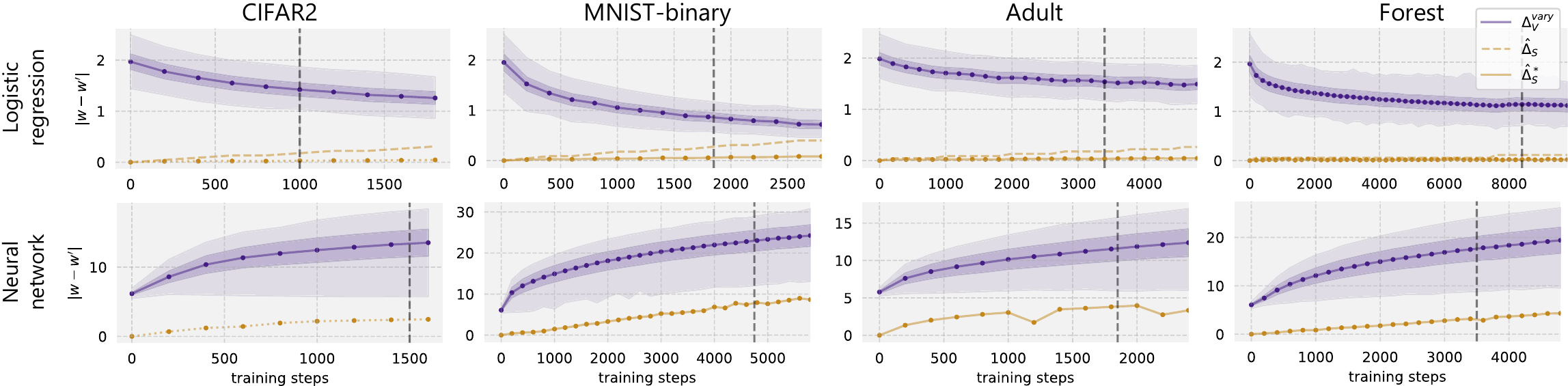}
    \caption{Value of $\|\mathbf{w} - \mathbf{w}'\|$ (distance between weights) as a function of training time, for pairs of experiments differing in data and not seed ($\wu$, gold bottom line), and pairs of experiments differing only in seed ($\var^{\text{vary}}$, purple top line).
    For logistic regression, the theoretical bound on the sensitivity ($\wubound$) is marked with a gold dashed line.
    The solid purple line (top) shows the average value of $\var^{\text{vary}}$ across different pairs of seeds, and the shaded areas show the full range (palest region) and 1 standard deviation.}
    \label{fig:versus_time}
\end{figure*}

We use our experimental set-up to explore this dependence and extend it to include the random seed.
We plot the theoretical sensitivity $\wubound$ (if available), estimated $\empbound$, and $\var^{\text{vary}}$ against the number of training steps $T$ in Figure~\ref{fig:versus_time}.

\begin{figure*}[h]
\centering
    \includegraphics[scale=0.46]{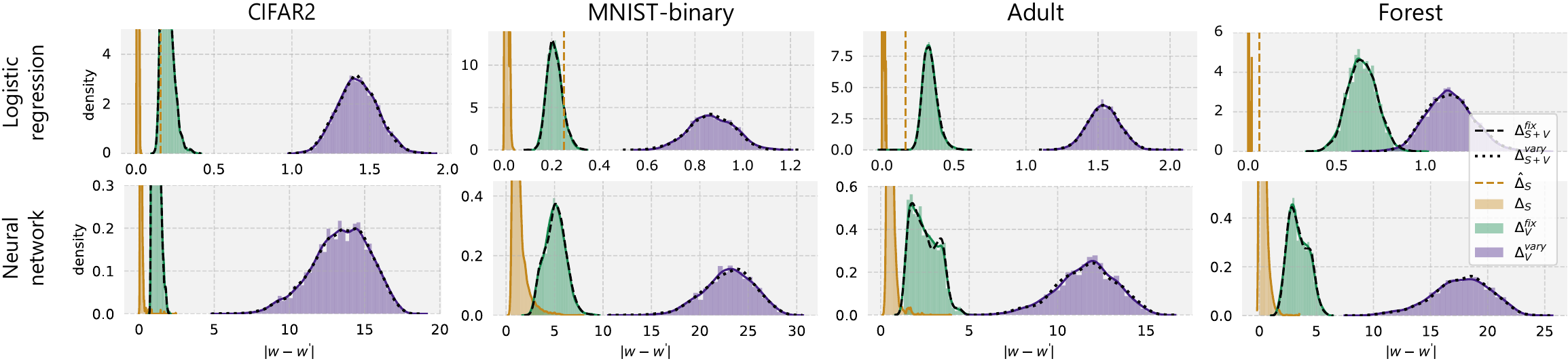}
    \caption{Distribution of $\|\mathbf{w} - \mathbf{w}'\|$ (distance between weights) across pairs of experiments.
    Left to right, curves show pairs differing in data ($\wu$), random seed with fixed model initialisation ($\var^{\text{fix}}$), and with variable initialisation ($\var^{\text{vary}}$).
    The change in $\mathbf{w}$ is dominated by the random seed, as evidenced by $\var$ tending to be much larger than $\wu$ and variation in both ($\wuvar$, denoted by dashed or dotted lines) overlapping with $\var$.
    The vertical dashed line is the theoretical bound of $\wu$ from~\cite{wu2017bolt} (not available for neural networks).}
    \label{fig:delta_histogram}
\end{figure*}

We see that pairs of models trained with different seeds (purple, top line) are much more dissimilar in parameter-space according to the $\ell_2$ norm, compared to setting where the seed is fixed (gold, bottom line).
For logistic regression (top row), the effect of the seed decreases during training as models converge to the unique global optimum.
We also see that the empirical sensitivity $\empbound$ (gold solid line) grows with $T$, but with a slope much lower than predicted by theory (gold dashed line), suggesting that the theoretical sensitivity bound $\wu$ is quite loose.
For neural networks (bottom row), the distance increases over time, indicating that models are converging to increasingly distant locations in parameter space.
The shaded region around the solid line of $\var^{\text{vary}}$ shows the full range of distances, indicating that some pairs of seeds produce models which are more or less similar.
We explore this distribution next.

\subsection{Distribution at convergence.}
By picking a $T$, we can visualise the full distribution of distances between models at this time - this is shown in Figure~\ref{fig:delta_histogram}.
We select $T$ as the convergence point, shown as vertical lines in Figure~\ref{fig:versus_time}.
The growth of $\empbound$ highlights the importance of converging early (if possible), both for generalisation as argued by~\cite{hardt2015train}, and -- as we argue, for privacy.

From Figure~\ref{fig:delta_histogram} we see that in almost all cases, $\var^{\text{vary}} > \wu$, indicating that changing the seed almost always has a larger impact than perturbing the dataset.
Fixing the initialisation ($\var^{\text{fix}}$) closes the gap to $\wu$, highlighting the importance of allowing model initialisation to vary with random seed.

For convex models, the theoretical bound $\wubound$ (vertical dashed line) is quite loose for all datasets, exceeding the largest observed value of $\wu$ by a factor between 3.2 and 6.5.
This suggests that even without accounting for intrinsic noise in SGD, existing approaches are likely over-estimating the sensitivity of SGD, and adding unnecessary noise for the desired privacy guarantees.
Investigation yielded the observation that the objective was rarely close to its Lipschitz bound, especially later in optimisation.

For neural networks, a longer tail in the distribution of data sensitivity indicates that in a small number of cases, a dataset perturbation can produce a change in the final model weights comparable to changing the random seed.
This may be caused by a small number of data points, suggesting that constraining to a data subspace may further decrease the sensitivity, as in the case of smooth sensitivity~\cite{nissim2007smooth}.
This behaviour may also be caused by the differing data point being encountered near the \emph{start} of training, a phenomenon also observed by~\cite{hardt2015train}.

\subsection{Projection of Weights with UMAP}
\begin{figure}
\centering
    \includegraphics[scale=0.75]{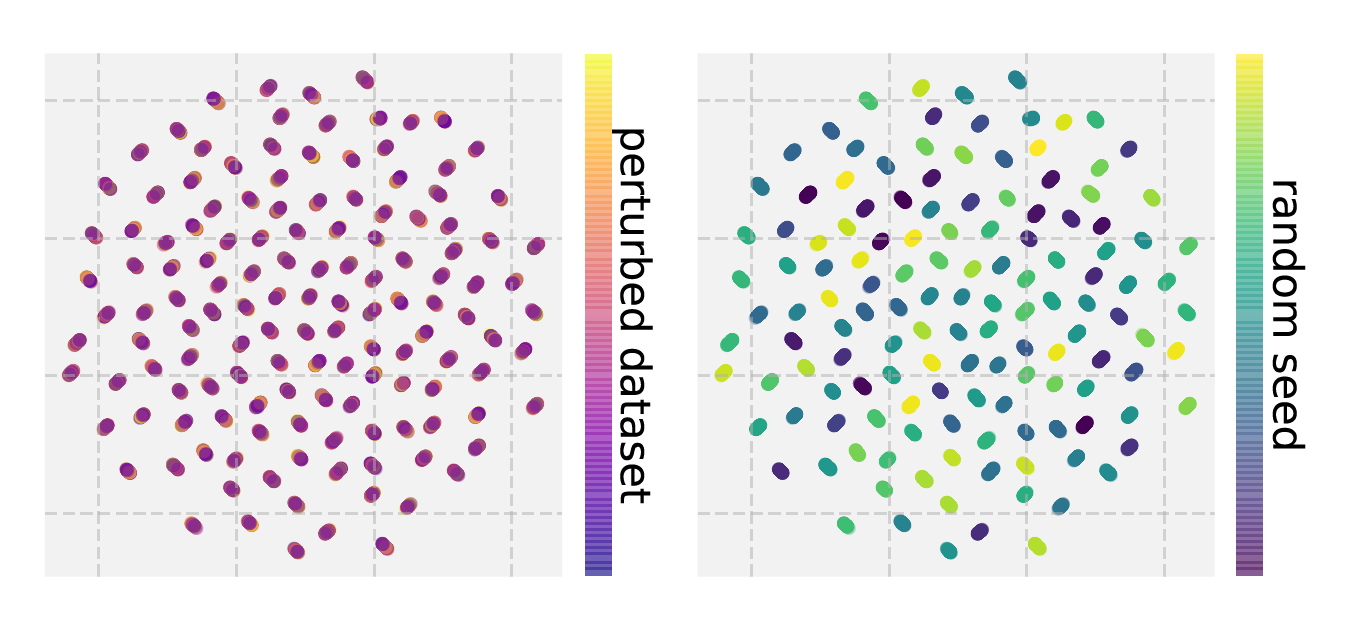}
    \caption{UMAP-embedded model weights, coloured by dataset (left) or random seed (right). Results shown are for logistic regression on \texttt{Forest} - other examples in Appendix~\ref{app:extended}.}
    \label{fig:umap}
\end{figure}

We can further visualise the impact of the random seed by projecting the model weights into a two-dimensional space using UMAP~\cite{mcinnes2018umap-software}.
This allows us to inspect the natural structure arising in the trained model weights at the convergence point $T$.
By colouring each point (trained model) by its underlying data instance (e.g. $S_i$), \emph{or} the seed used in training $r$ - shown in Figure~\ref{fig:umap} we can see UMAP has identified clusters which appear to separate \emph{seeds} very clearly (right), but \emph{not} data instances (left).

Performing a k-means clustering on the weight vectors confirms that models sharing the same random seed are mapped to the same cluster, and when the number of clusters equals or exceeds the number of seeds, each cluster contains at most one seed. Conversely, discovered clusters contained a variety of dataset instances.
We show results for logistic regression on \texttt{Forest} (chosen randomly), but find qualitatively identical results for all models and datasets (see Appendix~\ref{app:extended}).
Overall, we see clear evidence that the seed plays a larger role than dataset perturbations in the final weights of the model.

\section{What is the distribution of noise in SGD?}
\label{section:distribution}
We have shown that the seed has a significant impact on the weights of the model relative to data perturbations.
In this section, we ask what this impact looks like: specifically what is the resulting \emph{distribution} over model weights induced by the randomness in SGD?
Having this characterisation is key to analysing the extent to which noise in SGD could be leveraged for privacy, however as outlined in Section~\ref{section:related} theoretical results do not yet exist.

We use our experimental set-up to provide initial, \emph{empirical} evidence to this end.
We are particularly interested in whether SGD produces a distribution amenable to existing DP-mechanisms, such as Gaussian or Laplace.
To examine the distribution we collect more samples, running 800 additional seeds for each of the four benchmark datasets.\footnote{We selected $N = 800$ seeds through simulation that are sufficient to reject 50-dimensional non-normally-distributed variables, see Appendix~\ref{app:normality}.}
We then test the distributions against 1) multivariate Gaussian, 2) univariate Laplace (on weight marginals), and 3) for cases which could be multivariate Gaussian, we estimate the heaviness of the tails of the distribution ($\alpha$) as per~\cite{mohammadi2015estimating} (via ~\cite{Simsekli2019ATA}).

\subsection{Multivariate Gaussian}
We use a Henze-Zirkler (HZ) test~\cite{henze1990class} to test for multivariate Gaussian distributed data.
For \emph{convex objectives}, the p-values from the HZ test ranged between 0.14 and 0.25, indicating that the posterior distribution of the weights is \emph{not inconsistent} with a multivariate normal distribution for all datasets considered.

\begin{figure}[h]
\centering
    \includegraphics[scale=0.43]{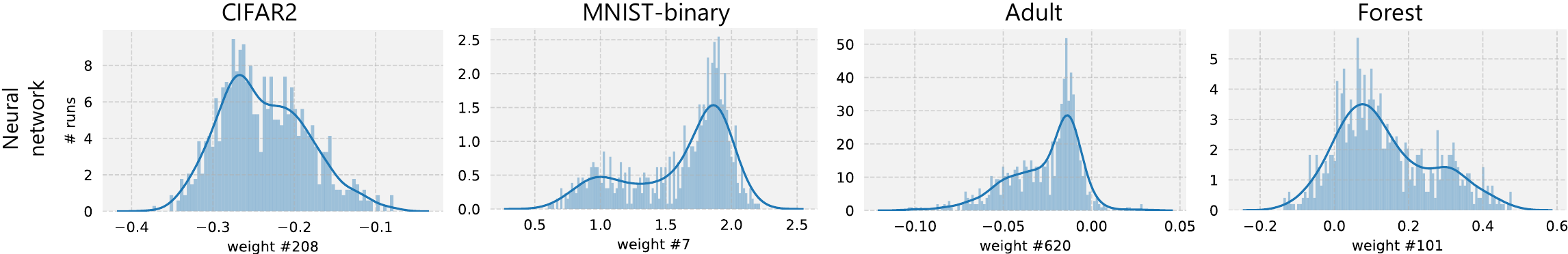}
    \caption{Examples of non-Gaussian weight posteriors (at convergence) for neural networks for each of the datasets.}
    \label{fig:non_normal_posteriors}
\end{figure}

For \emph{non-convex} objectives, we find that the posterior distribution of weights is \emph{not} consistent with a multivariate normal distribution.
Running the HZ test on random subsets of weights for computational reasons, the test rejects the null hypothesis in \emph{some} dimensions.
We can further confirm this by performing univariate tests of normality on each dimension independently using the Shapiro-Wilk test~\cite{razali2011power}, where we observe that some dimensions are not even univariate Gaussian, likely arising from the existence of multiple minima - we show examples in Figure~\ref{fig:non_normal_posteriors}.

The existence of non-normally-distributed parameters confirms that the joint cannot be multivariate Gaussian.

\subsection{Univariate Laplace}
We test individual dimensions of the weight distribution against a univariate Laplace distribution using the Anderson-Darling test.
The Laplace mechanism is a popular $(\epsilon, 0)-$differentially private mechanism, however we find the weight distribution of SGD, across the datasets and models considered, is not consistent with this distribution.
Accounting for multiple tests with the conservative Bonferroni correction, we find that between 38\% and 74\% of weights are \emph{not} Laplace, ruling out a joint Laplace distribution.

\subsection{Heaviness of tails}
The heaviness of the tails in SGD iterates has recently been demonstrated~\cite{gurbuzbalaban2020heavy}, with further implications for the theory of SGD.
We are in a position to test the `heaviness' of the tails of the weight distribution, using the estimator from ~\cite{mohammadi2015estimating}. This provides an estimate of the $\alpha$ parameter of an assumed-$\alpha$-stable distribution.
$\alpha = 1$ and $\alpha = 2$ correspond to Cauchy and Gaussian distributions respectively.

For the convex models, $\alpha$ ranges between 2.01 and 2.05, which further supports the similarity to a Gaussian.
This is also consistent with ~\cite{gurbuzbalaban2020heavy}, who report $\alpha \sim 2$ for batch sizes $\geq 20$.
We do not estimate $\alpha$ for neural networks as the estimator assumes a symmetric $\alpha$-stable distribution, which given previous results, we do not know to be the case.

\section{Estimating $\epsi$ for SGD on convex objectives with analogy to the Gaussian mechanism}
\label{section:epsilon}
Given the finding in the previous section that the posterior for convex objectives is \emph{consistent} (i.e. not inconsistent) with a Gaussian distribution, in this section we interpret SGD (on convex objectives) as a kind of Gaussian mechanism, where the Gaussian noise is introduced by the randomness in SGD.

We can relate the $\epsilon$ of a Gaussian-distributed query given $\delta$, its sensitivity $\Delta_2(f)$, and its variance $\sigma^2$ using Theorem~\ref{defn:gaussian}.
\begin{thm}{\bf From~\cite{Dwork2014TheAF}}
 Let $ f$ be a function that maps a dataset to a vector in $\mathbb{R}^d$.
Let $\epsilon \in (0, 1)$ be arbitrary. For $c^2 > 2 \ln \left(1.25/ \delta\right)$, adding Gaussian
noise sampled using the parameters $\sigma \geq c \Delta_2(f)/ \epsilon$ guarantees $(\epsilon, \delta)$-differential
privacy.
\label{defn:gaussian}
\end{thm}
Our objective is therefore to estimate (dataset-dependent) $\Delta_2(f)$ and $\sigma^2$ for (convex) SGD, to relate these to a potential $\epsilon$.
The sensitivity $\Delta_2(f)$ can be computed using either the theoretical bound (Equation~\ref{eqn:wubound}) or our sensitivity estimate (Equation~\ref{eqn:empirical_sensitivity}).

To estimate $\sigma^2$ we build on the previous section and \emph{assume}
\begin{equation}
A(S) \sim \mathcal{N}\left(\mathbf{w}_S ; \text{diag}\left(\sigma_1^2, ..., \sigma_d^2 \right)\right)
\label{eqn:gaussian}
\end{equation}
Where $A(S)$ refers to the distribution of values SGD will produce when run on dataset $S$, and $\mathbf{w}_S$ captures the dataset-dependent component.
Each weight is assumed independent but with its own variance $\sigma_j^2$.\footnote{In practice correlation between weights is likely, which could make use of non-diagonal multivariate Gaussian mechanisms~\cite{chanyaswad2018mvg}.}
The most conservative $\epsilon$ implied by the above expression is defined by the \emph{smallest} $\sigma_j$.
When estimated from data, we refer to this smallest $\sigma_j$ as $\sigmai$.
Estimation of $\sigmai$ from the experiment grid simply requires the standard deviation of each weight independently, after subtracting its dataset-specific mean.

Acknowledging therefore that this is not a true $\epsilon$ in the sense of $(\epsilon, \delta)$-differential privacy as it is estimated empirically, and that it is dependent on the underlying dataset $\mathcal{D}$ and model (implied), we call it $\epsi$.
It is defined as follows:
\begin{equation}
    \epsi = \frac{\sqrt{2 \log{1.25/\delta}} \wu}{\sigmai}\\
    \label{eqn:epsi}
\end{equation}
where $\wu$ is the sensitivity of SGD for that dataset and model, and $\sigmai$ is as described in the previous paragraph. We set $\delta = 1/N^2$ following convention~\cite{Dwork2014TheAF}.
We also consider $\epsi^*$ computed using the empirical sensitivity $\empbound$ instead of $\wu$.

As the Gaussian mechanism requires $\epsilon \in (0, 1)$, we interpret $\epsi$ as a way to capture the relationship between sensitivity and variability of SGD (for a given dataset).
Although the notation does not capture it, we assume an implicit model-dependence of $\epsi$ throughout.

\begin{algorithm}[h]
\begin{algorithmic}[]
    \REQUIRE Given neighbouring datasets $\mathcal{S} = \{S_a\}_a^{|S|}$, random seeds $\mathcal{R} = \{r_i\}_i^R$, SGD algorithm $A$ with batch size $B$ on convex loss, fixed learning rate $\eta$, number of epochs $k$, $\delta$, Lipschitz constant $L$.
    \FORALL{$S_a \in \mathcal{S}$, $r \in \mathcal{R}$}
            \STATE $\mathbf{w}_{r, a} \gets A(S_a; r)$ \COMMENT{Run SGD on $S_a$ with seed $r$}
    \ENDFOR   

\STATE{\hspace{-0.5cm}\textit{Compute sensitivity}}
	    \FORALL{$r \in \mathcal{R}$, $S_a, S_b \in \mathcal{S}$}
	            \STATE $\wu^{rab} \gets \|\mathbf{w}_{r, a} - \mathbf{w}_{r, b}\|$ \COMMENT{Pairwise sensitivity}
	    \ENDFOR
	    \STATE $\wubound \gets 2kL\eta/B$ \COMMENT{Theoretical bound}
	    \STATE $\empbound \gets \max_{r, a, b} \wu^{rab}$ \COMMENT{Empirical bound}
\STATE{\hspace{-0.5cm}\textit{Compute variance}}
	    \FORALL {$S_a \in \mathcal{S}$} 
	        \STATE{$\bar{\mathbf{w}}_a \gets \frac{1}{R}\sum_r \mathbf{w}_{r, a}$}
            \STATE{$\sigma_a^j \gets \text{stddev}\left(\mathbf{w}_{r, a}^j - \bar{\mathbf{w}}^j_a\right)$}
            \STATE{$\sigma_a \gets \text{min}_j \sigma_a^j$}
	    \ENDFOR
	    \STATE $\sigma_i \gets \text{min}_a^{|S|}\sigma_a$    \COMMENT{Variability}
    
\STATE{\hspace{-0.5cm}\textit{Compute epsilon}}
	    \STATE $c \gets \sqrt{2 \text{log}(1.25)/\delta} + 1 \times 10^{-5}$
	    \STATE $\epsilon_i \gets c\wubound/\sigma_i$ \COMMENT{Using theoretical sensitivity}
	    \STATE $\epsilon_i^{*} \gets c\empbound/\sigma_i$ \COMMENT{Using empirical sensitivity}
	    \RETURN $\epsilon_i, \epsilon_i^{*}$
    \end{algorithmic}
    \caption{ Estimating $\epsilon_i$ empirically}
    \label{algo:epsilon}

\end{algorithm}

\begin{table*}[t]
\centering
        \begin{tabular}{l|c|c|c|c}
        & \texttt{CIFAR2}   & \texttt{MNIST-}   & \texttt{Adult} & \texttt{Forest} \\
        &                   & \texttt{binary}          &                & \\\hline
        $\delta$                 & $1.23 \textsf{x} 10^{-8}$ & $9.25 \textsf{x} 10^{-9}$ & $1.16 \textsf{x} 10^{-9}$ & $6.97 \textsf{x} 10^{-12}$ \\ \hline
        $\wubound$               & 0.157 & 0.252 & 0.164 & 0.063\\ 
        $\empbound$              & 0.030 & 0.059 & 0.036 & 0.020\\
        $\sigmai$                & \textbf{0.096} -- 0.140 & \textbf{0.021} -- 0.136 & \textbf{0.053} -- 0.148 & \textbf{0.072} -- 0.195 \\
        $\epsi$                  & 6.824 -- \textbf{9.933} & 11.662 -- \textbf{73.768} & 7.156 -- \textbf{19.978} & 2.316 -- \textbf{6.313} \\
        $\epsi^*$                  & 1.298 -- \textbf{1.890} & 2.791 -- \textbf{17.653} & 1.480 -- \textbf{4.133} & 0.722 -- \textbf{1.969} \\\hline
        \end{tabular}
            \caption{Theoretical sensitivity ($\wubound$), empirical sensitivity ($\empbound$), privacy parameter $\delta$ ($1/N^2$), intrinsic variability $\sigmai$ accounting for variable initialisation, intrinsic $\epsi$, and intrinsic $\epsi^*$ computed using the empirical bound. $\sigmai$ is denoted in bold-face, however we include the range of observed weight-specific standard deviations. Similarly for $\epsi$ and $\epsi^*$ we denote the proper value (using $\sigmai$) in bold, with ranges indicating weight-specific values.}

    \label{table:sigmas}
\end{table*}
The whole procedure is outlined in Algorithm~\ref{algo:epsilon}. We run this procedure on our grid of trained models, and report the findings in Table~\ref{table:sigmas}. Specifically, we report values for $\delta$, sensitivities both theoretical and empirical $\wu$ and $\empbound$, variance $\sigmai$, and resulting $\epsi$ for each of the four datasets, all for logistic regression.

For completeness, we include the range of all observed (empirical) weight-specific $\sigma_j$ ($\sigmai$ is the \emph{smallest} of these).
This results in ranges for $\epsi$ and $\epsi^*$.
These other values would arise in the case that some weights were excluded from the privacy analysis, perhaps because they only depend on non-private parts of the data.

Reflecting the looseness of the theoretical bound $\wu$ relative to the empirical sensitivity $\empbound$, we see that intrinsic $\epsi$ greatly exceeds $\epsi^*$.
This indicates that even without accounting for the intrinsic randomness of SGD, performance of private models using output perturbation could likely be improved by simply tightening the sensitivity bound.

We also note the large variation in `$\epsi$' arising from weight-specific variability.
This is most pronounced for \texttt{MNIST-binary}, where a small number low-variance weights push up the overall $\epsi$.
We also experimented with computing \emph{per-weight} sensitivity but found these low-variance weights were \emph{not} necessarily matched with low sensitivity. Designing methods to minimise the existence of low-variance weights is an interesting question for further research.
The fast convergence on \texttt{CIFAR2} benefits both its sensitivity and variability, leading to low $\epsi$, supporting the use of transfer from public datasets for privacy-preserving machine learning~\cite{papernot2019making}.
The marked variation between the four datasets further highlights the need to account for dataset factors in estimating both sensitivity and variability.

Finally, we see that the values of $\epsi$ range from 1.89 (\texttt{CIFAR2}, with empirical sensitivity bound) to 73.77 (\texttt{MNIST-binary}, with theoretical sensitivity). This lower range is comparable to values of (real) $\epsilon$ used in building differentially-private variants of SGD~\cite{jayaraman2019evaluating}\footnote{Apple uses an $\epsilon$ of 2, 4, or 8: \url{https://www.apple.com/privacy/docs/Differential_Privacy_Overview.pdf}, and the US Census in 2020 used $\epsilon = 19.61$ globally, with $\epsilon$ = 2.47 for housing unit data: \url{https://www.census.gov/newsroom/press-releases/2021/2020-census-key-parameters.html}}.
While this does \emph{not} constitute a statement that SGD is itself privacy-preserving, it provides strong empirical evidence that - subject to assumptions - the randomness in (convex) SGD is not \emph{trivial} in the sense of privacy, and should be further studied.

\subsection{Did we run enough experiments?}
\label{section:enough_experiments}
\begin{figure*}[h]
\centering
    \includegraphics[width=\textwidth]{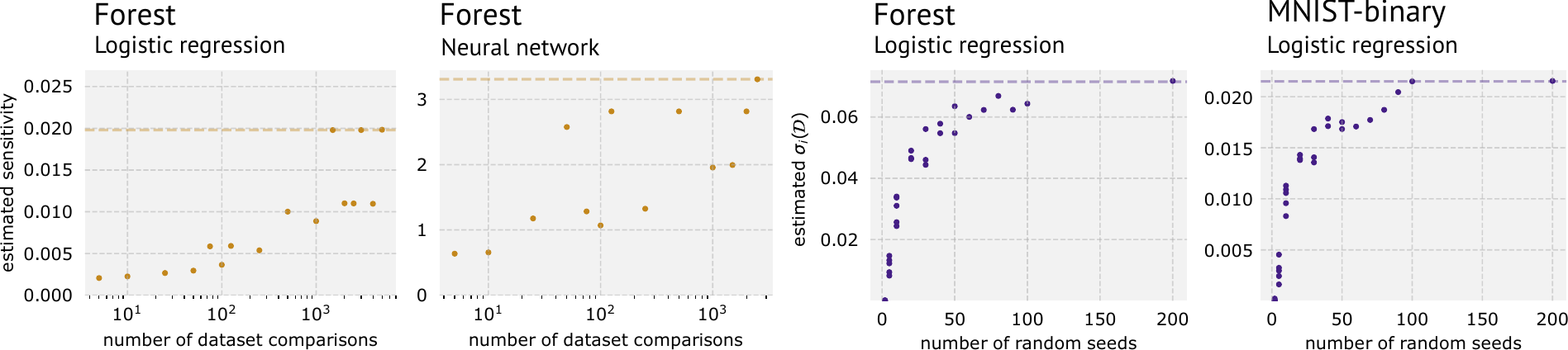}
    \caption{We explore how increasing the amount of data used to estimate $\empbound$ and $\sigmai$ change the estimates. The left two panels show $\empbound$ as a function of the number of pairs of neighbouring datasets (note log scale on $x$-axis). The right two panels show $\sigmai$ as a function of the number of random seeds. In both cases, dashed lines show the values used elsewhere in the paper. Results are shown for a subset of datasets and models, see Appendix~\ref{app:extended} for complete results. Note that we do not estimate $\sigmai$ for neural networks as their posterior is non-Gaussian.}
    \label{fig:consistency}
\end{figure*}

We have explored only a subset of the possible combinations of dataset perturbations and random seeds for each of our data sources, which may introduce uncertainty in our estimates of $\empbound$ and $\sigmai$.
To test this, in Figure~\ref{fig:consistency} we visualise how the estimates of $\empbound$ and $\sigmai$ change as we use more data (that is, include more experiments), for \texttt{Forest} (other datasets in Appendix Figures~\ref{fig:consistency_empbound} and \ref{fig:consistency_sigma}).

As we can see, as the number of experiments used to estimate the values increases, our estimates tend towards a fixed value, suggesting that more experiments would not \emph{substantially} alter the findings.
We see that we are likely under-estimating the sensitivity ($\empbound$), which is a natural consequence of it being the maximum of an unknown distribution.

In Table~\ref{table:sigma_is_stable} we show the range of values we estimate for $\sigmai$ using different dataset instances $\{S_i\}$, for all four benchmark datasets.
In practice we take the minimum value observed for each dataset as $\sigmai$, however Table~\ref{table:sigma_is_stable} indicates that the distribution of $\sigmai$ values is very narrow, and the minimum is very similar to a point estimate from a single dataset.
This result indicates that for the purpose of estimating $\sigmai$, it is likely unnecessary to run a `full' grid of seed and dataset pairs, and one could rather focus computational effort on running more seeds.

Overall the results in this section indicate that although we have only run a small fraction of the possible experiments, we would not expect our estimates to change greatly with more experiments.

\begin{table}[h]
\centering
    \begin{tabular}{r|l|l|l|l}
                        & \texttt{CIFAR2} & \texttt{MNIST-binary} & \texttt{Adult} & \texttt{Forest} \\\hline
            \# datasets       &   98 &       99     &       98      &       75      \\
            min         &   0.096\textbf{7}  & 0.02\textbf{1}     &   0.053\textbf{0}  &  0.07\textbf{1}\\  
            max         &   0.096\textbf{8}  & 0.02\textbf{2}     &   0.053\textbf{2} &   0.07\textbf{3}\\\hline
    \end{tabular}
    \caption{The estimate of $\sigmai$ is stable across dataset instances $\{S_i\}$ across all four datasets (for logistic regression; we don't compute $\sigmai$ for neural networks).
    We report the minimum and maximum observed variability estimated from each dataset instance across many dataset instances, retaining digits until the first difference (shown in bold). Reporting these extreme statistics adequately capture the narrowness of the distribution.}
    \label{table:sigma_is_stable}
\end{table}

\section{Estimating $\epsilon$ for SGD on non-convex objectives using membership inference attacks}
\label{section:mi}

As shown in Figure~\ref{fig:non_normal_posteriors}, the weight posterior for non-convex models is not Gaussian and hence we cannot use Algorithm~\ref{algo:epsilon} to compute the inherent empirical epsilon. Recent research has demonstrated that epsilon lower bound can be empirically estimated using membership inference (MI) attacks~\cite{jagielski2020auditing,nasr2021adversary}. Hence, we use MI attack as a metric to compute the inherent epsilon and understand the effect of using a random seed in a non-convex setting.
Note that, it has been shown that MI attack provides a loose lower bound estimate for the epsilon guarantees of the model~\cite{jagielski2020auditing}. However, our goal is to understand the effect on the privacy guarantees of the model due to the change in random seed i.e., relative change in epsilon and not focus on the absolute epsilon values estimated using the attack.

We use the method mentioned in Yeom et al.~\cite{Yeom2018PrivacyRI} to compute our epsilon values from MI attack advantage. We observe that the small convex binary-classification models (Section~\ref{section:empirical_study}) do not contain sufficient signal to perpetrate MI attacks which is also shown in prior work~\cite{shokri2017membership}. Hence, we estimate the empirical epsilon {\em only} on sufficiently large non-convex models. Following Yeom et al.~\cite{Yeom2018PrivacyRI} for CIFAR-10, we train a modified VGG network. This model set-up includes a size parameter $s$, which we set to $s=2^4$ as it balances model size with attack advantage. The full architecture is described in Appendix~\ref{section:vgg}. The model has 82,554 parameters in total. As in~\cite{Yeom2018PrivacyRI}, we train on a subset of 15,000 examples from the full (10-class) CIFAR-10 dataset. As for earlier experiments, we train using SGD with a fixed step-size.

\begin{table*}[t!]
    \centering
    \begin{tabular}{l|lll|lll|l|lc}
         & Accuracy &  & &  $\epsilon$ &   &  &  &  &  \\
    T    & Same & Different & $\Delta$ & Same & Different & $\Delta$ & p-value    & \# M    & $R_\textrm{test}$/$R_\textrm{train}$ \\\hline
    2000* & 0.5263   & 0.5262    &  0.0000918 &  0.1055   & 0.1051    &0.000366  & 0.0325687  & 1500 & 1.13                      \\
    4000  & 0.5909   & 0.5905    & 0.0004042 &  0.3677   & 0.3662    & 0.001591  & 0.0002862  & 1000 & 1.86                      \\
    8000  & 0.6593   & 0.6586    &  0.0007116 & 0.6610   & 0.6580    & 0.00294   & 1.3734e-05 & 1000 & 5.23                    \\
    \end{tabular}

     \caption{Results from membership inference against a CNN trained with vanilla SGD on CIFAR-10. $T$ is number of training steps after which the CNN is attacked. `Same' and `Different' refer to whether the adversary chose the loss threshold using a model with the same or different seed to the target model. The p-value is from a one-sided paired t-test between $\epsilon$ (Same) and $\epsilon$ (Different). \#M: number of attacks performed for experiment. $R$ refers to the model's loss (cross-entropy) on either training or heldout (test) data. As these models were trained without differential privacy, the theoretical $\epsilon$ is $\infty$.}
         \label{table:mi_attack}
\end{table*}

As in Yeom et al.~\cite{Yeom2018PrivacyRI}, our MI attack uses a loss-based threshold to distinguish between members and non-members. To explicitly test for the impact of the random seed, we consider two types of adversarial setting:
\begin{enumerate}
    \item When the threshold is selected from the same model as the target model i.e., adversary has knowledge of the random seed used to train the model
    \item When the threshold is selected from a model trained using a different seed.
\end{enumerate}
The threshold is computed as the mean of a subset of training samples different than the one used to attack the target model. Similar to Algorithm 6 in ~\cite{jagielski2020auditing}, we estimate epsilon based on adversary's advantage for the setting when target model is trained using the same and different seed. Analogously to prior work of `instantiating' adversaries~\cite{nasr2021adversary}, by comparing the performance of these two adversaries we can directly quantify the impact of having knowledge of the random seed. 

We run the attack on at least 1000 different models and report average values for $\epsilon$ and accuracy. We compute the difference in the two setting to understand the effect size of the random seed on the epsilon and report the corresponding p-value for significance, arising from a one-sided paired t-test. We also compute the ratio of train vs validation loss similar to Yeom et al. Table~\ref{table:mi_attack} shows results for increasing values of $T$, starting at the model's convergence point of $T=2000$. As in prior work, the absolute performance of the MI attack increases as the model overfits; at $T=4000$ our results mirror those of \cite{Yeom2018PrivacyRI} (they do not report later timepoints for this setting). In general, the values of $\epsilon$ estimated by the MI attack are underestimates, as highlighted by ~\cite{jagielski2020auditing}, especially as we are attacking a model trained only with `vanilla' SGD, corresponding to a \emph{theoretical} $\epsilon$ of $\infty$ (ignoring any inherent randomness). However, our focus is on the \emph{change} in the MI attack performance due to hiding of the seed. Looking then at the deltas, we find a significant reduction in the performance of the adversary according to the paired t-test, using a significance level of 0.05. The size of this effect becomes more marked with increasing $T$. Hiding the random seed from the adversary results in an effective decrease of $\epsilon$ of 0.003 at $T=8000$.

Overall these results show that concealing the random seed from the adversary increases the effective $\epsilon$ afforded by SGD as measured by a membership inference attack.

\section{Implications and Conclusion}
We have taken first steps towards examining the inherent noise of SGD from the perspective of privacy. By performing a large scale empirical study, we have demonstrated:
\begin{enumerate}
    \item The seed used to train a model plays an important role in the final weights. Specifically:
    \begin{enumerate}
        \item Models trained with the same seed, even on neighbouring datasets, have more similar weights, than those trained on the same data, with differing seed.
        \item Hiding the seed used to train a model from an adversary results in significantly worse membership inference performance, albeit with small effect size.
    \end{enumerate}
    \item Theoretical bounds on the sensitivity of SGD for convex objectives are loose by up to a factor of 6.5.
    \item The distribution of weights resulting from SGD on convex objectives can be modelled with a multivariate Gaussian distribution. Under the assumption of a diagonal multivariate Gaussian weight distribution, we can compute an `intrinsic' $\epsilon$ by analogy to the Gaussian mechanism, and find values between 1.9 and 17.6 across four datasets.
    \item Non-convex objectives \emph{empirically} have weight distributions that do not follow `simple' distributions such as Gaussian or Laplace.
\end{enumerate}

These findings have the following implications:
\begin{enumerate}
    \item Inherent stochasticity during model training (such as that arising in SGD) has been under-explored. Assuming the seed can remain private during training empirically weakens adversaries, and is not an unreasonable assumption. Exploiting, or perhaps enhancing, stochasticity during training may allow for improved differentially-private training algorithms. This complements findings from \cite{nasr2019comprehensive}, who highlighted the need to place additional assumptions on adversaries to achieve better privacy guarantees.
    \item There remains an opportunity to tighten the theoretical bounds on the sensitivity of SGD, which would reduce the scale of noise required for output perturbation as used in \cite{wu2017bolt} and provide improved private model performance.
    \item Given non-convex objectives have complex posterior distributions, further theoretical work is required to characterise them to support the development of private algorithms.
    \item However, these results do \emph{not} indicate that the inherent noise in SGD is sufficient on its own for privacy. These findings should be used to further tighten existing bounds.
\end{enumerate}

We hope that this work inspires further work into the properties of stochastic gradient descent and other stochastic learning algorithms, and highlights the benefit of empirical analysis for supporting ongoing theoretical work.
All code for conducting these experiments is available at \url{https://github.com/microsoft/intrinsic-private-sgd}.

\bibliographystyle{plain}
\bibliography{paper}

\appendix
\section{Further details on the sensitivity of SGD for convex functions}
\label{app:sensivity}
In this section we provide an expanded explanation of the theoretical sensitivity bound and provide some initial analyses into the role of \emph{randomness} in the output of SGD.
These findings are restricted to a particular class of loss functions:

\textbf{Assumptions.}
Let $\mathcal W \subseteq \mathbb R^p$ be the hypothesis space, and $\loss: \mathcal W \mapsto \mathbb R$ the loss function.
We assume that $\forall u, v \in \mathcal W$:

\begin{itemize}
\item $\loss$ is convex;
i.e., $\loss(u) \ge \loss(v) + \langle \nabla \mathcal(v), u - v \rangle$

\item $\loss$ is $L$-Lipschitz 
i.e., $\| \loss(u) - \loss(v) \| \le L \| u - v \|$

\item $\loss$ is $\beta$-smooth;
i.e., $\| \nabla \loss(u) - \nabla \loss(v) \| \le \beta \| u - v \|$
\end{itemize}

 We present the results for the sensitivity of SGD due to a change input data point as provided by ~\cite{wu2017bolt}:
 \begin{thm}{\bf From ~\cite{wu2017bolt}.}
Let A denote the SGD algorithm using $r$ as the random seed then the upper bound for sensitivity for $k$-passes of SGD is given by
\begin{equation*}
\wubound = \max_r \| A(r; S) - A (r; S')  \| \le  2kL\eta
\end{equation*}
\label{theorem:wubound}
\end{thm}
Here, $\wubound$ gives the maximum difference in the model parameters due to the presence or absence of a single input sample. 
Their results rely on the {\em boundedness} and {\em expansiveness} properties for the gradient update rule ($G$) of SGD as proposed by \cite{hardt2015train}:

\begin{property}({\bf Boundedness of G.})
For a loss function that is $L$-Lipschitz and learning rate $\eta$, the gradient update of SGD is $\eta L$ bounded i.e.,
\begin{equation*}
 \sup_{w \in \mathcal W} \| G(w) - w\| \le \eta L
 \end{equation*}
 \label{eq:boundedness}
\end{property}

\begin{property}({\bf Expansiveness of G.})
For a loss function that is $\beta$-smooth, and $\eta \le 2/\beta$, then the gradient update of SGD is $1$-expansive i.e.
\begin{equation*}
\sup_{w, w'} \frac{\|G(w) - G(w') \|}{\|w - w'\|} \le 1 
\end{equation*}
 \label{eq:expansiveness}
\end{property}
We refer interested readers to the original paper for a formal proof~\cite{wu2017bolt}, and provide a brief intuition here for achieving the bound:
For a single pass of SGD over neighbouring datasets $S$ and $S'$ with a {\em fixed  initialization} and {\em fixed sampling strategy}, the two executions $G$ and $G'$ will differ only at a single step --- when the differing sample gets selected.
In that case, from the above boundedness property, we have that $G(w) - G'(w') \le 2L\eta $. For all the other steps, the samples selected are exactly same and hence the $1$-expansiveness property applies.
Therefore, after $k$-passes of SGD over the dataset, the difference in the model parameters will have an upper bound of $2kL\eta$.  
When trained using a batch size of $B$, the sensitivity bound can be reduced by a factor of $B$ i.e., $\wubound  \le  2kL\eta / B$ as only a single element in a given batch can be differing.

\section{Implementation and training details}
\label{app:implementation}

\subsection{Training and hyper-parameters}
Model parameters were initialised according to the `glorot-uniform' setting in Keras, which is default.
Biases were initialised to zero.
We used a batch size of 32 for all datasets, except for \texttt{Forest} where $B = 50$, to replicate~\cite{wu2017bolt}.
This validation set was used to select hyper-parameters and convergence point, based on validation performance failing to improve three times in a row, or by visual assessment of loss curves.
Validation performance was computed every 50 or 100 batches.

Table~\ref{table:hyperparameters} shows the learning rate ($\eta$), convergence point ($T$), number of experiments ($E$), number of model parameters ($P$), and hidden size (for neural networks) for each model and dataset.
The batch size, learning rate, and convergence point appear in the theoretical sensitivity bound (Theorem~\ref{theorem:wubound}) and these parameters likely influence the variability of the resulting weight distribution.
For the purpose of studying the seed alone, we consider such hyper-parameters fixed.

\begin{table}

\centering
    \begin{tabular}{l| l| l| l | l| l | l}
        Model & Dataset & $\eta$ & $T$ & $E$ & $P$ & Hidden size\\\hline
        \texttt{LogReg} & \texttt{CIFAR2} &  0.5 & 1000 & 9900 & 65 & -\\
        & \texttt{MNIST-binary} &  0.5 & 1900 & 19600 & 51 & -\\
        & \texttt{Adult} & 0.5 & 3400  & 25700 & 101 & -\\
        & \texttt{Forest} & 1.0 & 8400 & 20693 & 50 & -\\\hline
        \texttt{NN} & \texttt{CIFAR2} & 0.25 & 1500 & 9900 & 661 & 10\\
        & \texttt{MNIST-binary} & 0.5 & 4750 & 9800 & 521 & 10\\
        & \texttt{Adult} & 0.5 & 1850  & 11250 & 817 & 8 \\
        & \texttt{Forest} & 0.5 & 3500 & 11247 & 511  & 10\\
    \end{tabular}
       \caption{Training and model hyper-parameters. 
    $\eta$ is the fixed learning rate. $T$ is the number of training steps (we take this as the convergence point of the model).
    $E$ is the number of experiments performed, and $P$ is the number of parameters in the model. 
    \texttt{LogReg} refers to logistic regression models, and \texttt{NN} to neural networks.}
    \label{table:hyperparameters}
\end{table}

\subsection{Fine-tuning on CIFAR2}
We `project' CIFAR-2 down from its original (32, 32, 3) shape to $d = 64$ using a ResNet-56 which was pre-trained on CIFAR-100. We use the pre-trained model provided here: \url{https://github.com/chenyaofo/CIFAR-pretrained-models}.
We produce the `embedding' by removing the final, fully-connected layer of the ResNet-56. Training a logistic regression model on this ResNet-embedded data is then equivalent to fine-tuning where all but the final layer(s) of the model is frozen, and where a scaling layer (such that $\|\mathbf{x}\| \leq 1$) is introduced.
Training a neural network with a single hidden layer is equivalent to `fine-tuning' a variant of the ResNet-56 with an additional fully-connected layer before its final layer, as well as the scaling layer.
We pre-process the data (rather than explicitly fine-tune with frozen layers) to avoid having to run the large ResNet-56 many times.

\subsection{VGG on CIFAR-10}
\label{section:vgg}
We have implemented the architecture as described in~\cite{Yeom2018PrivacyRI}. The layers are listed in Table~\ref{table:vgg}.
\begin{table}[h]
\centering
    \begin{tabular}{|l | l | l|}\hline\\
    Layer & Size & Activation\\\hline\hline
    Convolution & Kernel size 3, $2^4$ filters & ReLU \\
    Convolution & Kernel size 3, $2^4$ filters & ReLU \\
    Max pooling & Pool size 2 & \\
    Convolution & Kernel size 3, $2^5$ filters & ReLU \\
    Convolution & Kernel size 3, $2^5$ filters & ReLU \\
    Max pooling & Pool size 2 & \\
    Flattening &    & \\
    Fully-connected & Output size $2^5$ & ReLU \\
    Fully-connected & Output size 10 & Softmax \\\hline
    \end{tabular}
        \caption{Architecture of CNN used for CIFAR-10. Total number of parameters is 82,554.}
    \label{table:vgg}
\end{table}

For this model we also train with a fixed learning rate of 0.1 and batch size of 32. We train for a maximum of 20 epochs, on a subset of 15,000 training samples. This model achieves a test-set accuracy of about 50\% for ten-class classification at $T=2000$. This is far from state-of-the-art on CIFAR-10, however as we use a subset of the data and `vanilla' SGD, this is not unexpected.

\subsection{Frameworks}
Experiments were implemented in Keras\cite{chollet2015keras}, TensorFlow\cite{tensorflow2015-whitepaper}, with PyTorch\cite{NEURIPS2019_9015} for the pre-trained CIFAR-100 model.
Wrangling of results and generation of figures relied on scikit-learn~\cite{scikit-learn}, pandas~\cite{mckinney-proc-scipy-2010}, matplotlib~\cite{Hunter:2007}, and seaborn~\cite{michael_waskom_2018_1313201}.

Experiments were run on a standard Azure Data Science Virtual Machine.
Code is available at \url{https://github.com/microsoft/intrinsic-private-sgd}.

\section{Normality testing and HZ test}
\label{app:normality}
In this section we provide more details on how we tested the posterior distribution of weights.
We use the implementation of the Henze-Zirkler (HZ) test~\cite{henze1990class} from Pingouin~\cite{Vallat2018PingouinSI}.

\subsection{Stability of HZ.}
This test is numerically unstable for data with dimension larger than approximately 55.
For models parameter count $d > 55$, we select $2 (d // 55 + 1)$ random subsets of parameters to test jointly.
Since all marginals of a multivariate normal distribution are themselves (multivariate) normal, we consider a rejection of the null hypothesis in any of these subsets to constitute an overall rejection, and therefore report the \emph{minimum} p-value across subsets.
We note that the \emph{average} p-value computed in this fashion was often relatively large, indicating that only a subset of dimensions were failing to be jointly normal.

\subsection{Power of HZ.}
\begin{figure}[h]
\centering
    \includegraphics[scale=0.45]{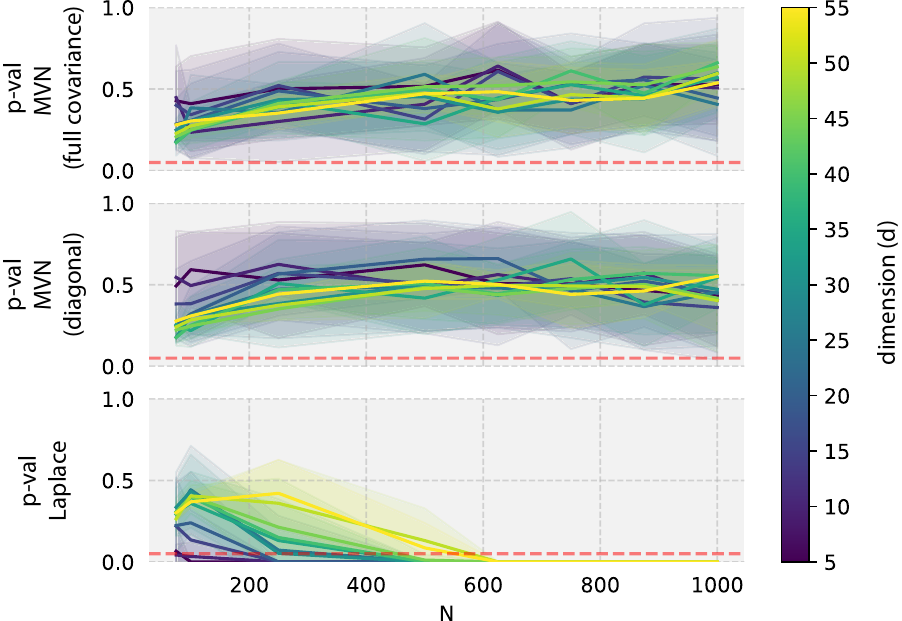}
    \caption{We show the p-value (y-axis) of the Henze-Zirkler test as a function of the number of samples ($N$, x-axis) and the dimensionality of the vector ($d$, colour).
    The traditional $p = 0.05$ line is marked in dashed red.
    Solid lines show averages across 10 replicates, and shaded areas show $\pm$ standard deviation.
    The first two rows correspond to tests performed on data that \emph{is} multivariate normal, with either full or diagonal covariance.
    The last row corresponds to Laplace-distributed data, indicating the test is powerful enough to reject this non-normal data for $N > 600$.}
    \label{fig:hz_test_power}
\end{figure}

We perform a simulation study on the HZ test to determine how many random seeds will be needed to identify a non-Gaussian weight posterior.
We generate ten replicates of data of shape $(N, d)$ from either a true multivariate normal distribution, with either full covariance (sampled from an Inverse Wishart prior) or diagonal covariance, as well as Laplace-distributed data.
We expect the HZ test to produce a `high' p-value for normal data (as it will not reject the null hypothesis), and low p-values for the Laplace data.
The result of this simulation is shown in Figure~\ref{fig:hz_test_power}, from which we conclude that $N = 800$ seeds should provide reasonable power.

\subsection{Univariate normal tests.}
\begin{figure}[h]
\centering
    \includegraphics[scale=0.65]{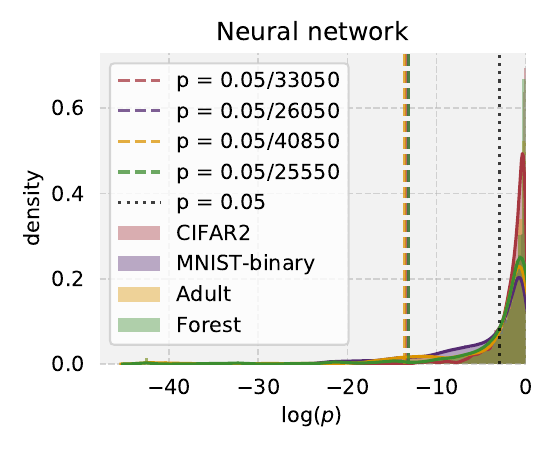}
    \caption{While most model parameters have marginal distributions consistent with a Gaussian by Shapiro-Wilk~\cite{razali2011power}, some dimensions do not, indicating the joint is not multivariate Gaussian. Density in this plot is over each parameter from each model, across 50 experiments.}
    \label{fig:pval_histogram_nonconvex}
\end{figure}

We use the Shapiro-Wilk~\cite{razali2011power} test of normality on individual weight distributions - see Figure~\ref{fig:pval_histogram_nonconvex} for the distribution of p-values across the four datasets for neural networks.

Vertical lines in the plot indicate the `standard' $p = 0.05$ cutoff, as well as thresholds corrected for multiple hypothesis testing.
For each model/dataset and each parameter, we compute the p-value using all the seeds from 50 dataset variants.
Using the Bonferroni correction we then have $p = 0.05/M$, where $M$ is the number of hypotheses, in our case this is the number of model parameters times the number of experiments, so $M = 50P$.

This indicates that \emph{most} parameters are \emph{usually} consistent with a univariate normal distribution, but violations exist, indicating the joint cannot be multivariate normal.

\section{Extended results}
\label{app:extended}
\subsection{UMAP}
\begin{figure*}
    \begin{minipage}{1\textwidth}
    \includegraphics[scale=0.72]{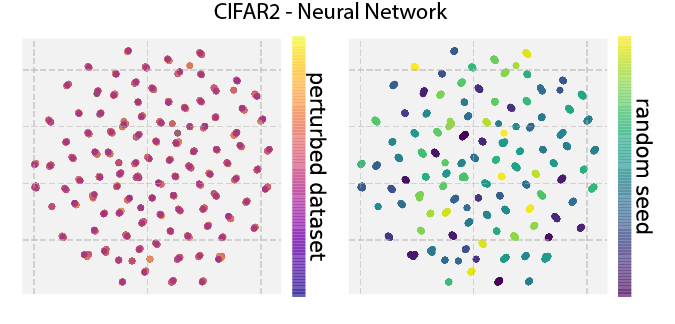}
    \includegraphics[scale=0.72]{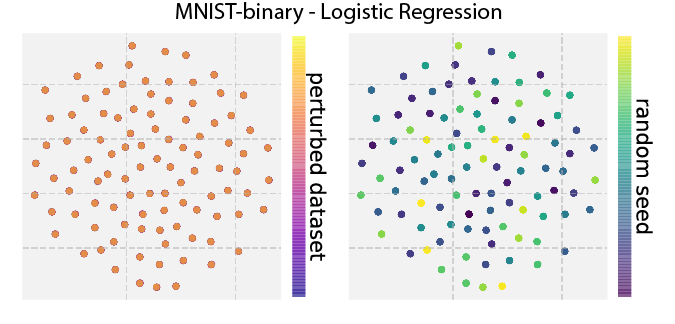}
    \end{minipage}
    \begin{minipage}{1\textwidth}
    \includegraphics[scale=0.72]{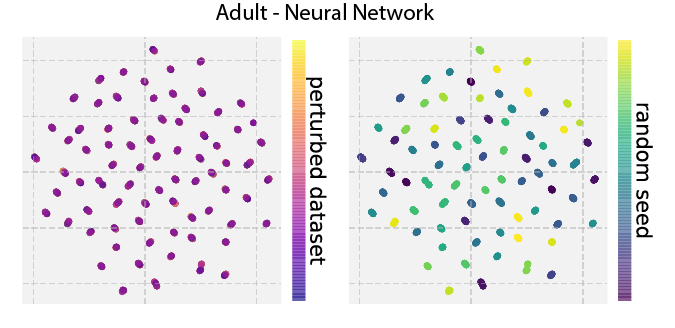}
    \includegraphics[scale=0.72]{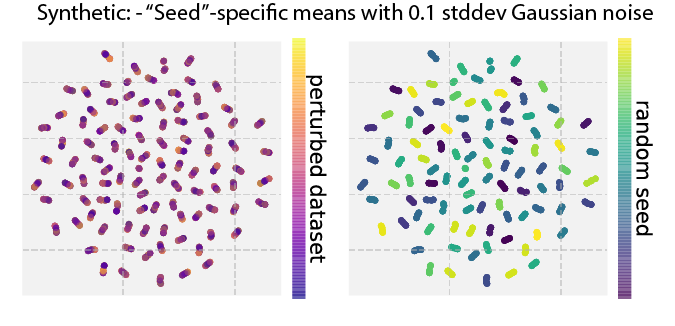}
    \end{minipage}
    \caption{Examples of UMAP embedding of weights at convergence for more datasets and a synthetic example.}
    \label{fig:more_umap}
\end{figure*}

This section includes further examples using UMAP to visualise model weights from different experimental conditions (Figure~\ref{fig:more_umap}).
We use the implementation of UMAP in \texttt{umap-learn}~\cite{mcinnes2018umap-software}. We experimented with different parameter choices (minimum distance, number of neighbours, metric) but did not see significant variation in the observation that groups `clustered' by seed much more clearly than by dataset instance.

We found the pattern surprisingly robust. Binarising the weights and using Hamming distance as the metric reproduced the pattern, as did performing UMAP on weights after a very small number of batches. In the latter case this is not surprising, as the seed impacts \emph{each} batch, while the differing data sample will only appear once per epoch, so the seed naturally has an immediate impact.
We confirmed that shuffling the weights for each model independently destroyed the effect (models are not a bag of weights).
We were also able to generate a similar pattern by running UMAP on synthetic data where each example consists of Gaussian noise around a seed-specific centre - the result with standard deviation 0.1 is shown in the last panel of Figure~\ref{fig:more_umap}.

\subsection{Did we run enough experiments?}
\begin{figure*}[h]
    \begin{minipage}{1\textwidth}
        \centering
        \includegraphics[scale=0.55]{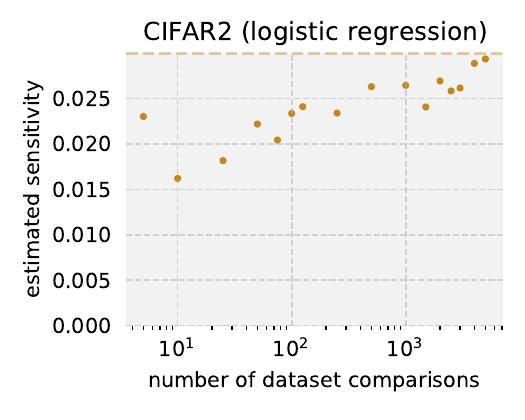}
        \includegraphics[scale=0.55]{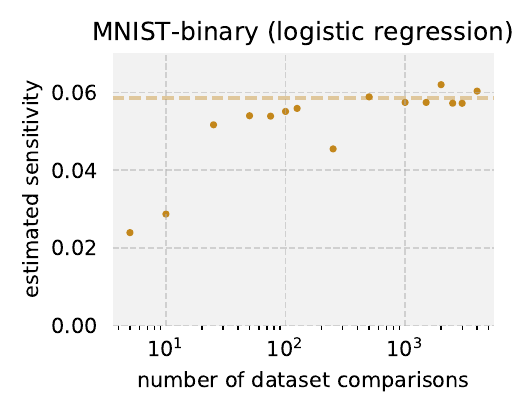}
        \includegraphics[scale=0.55]{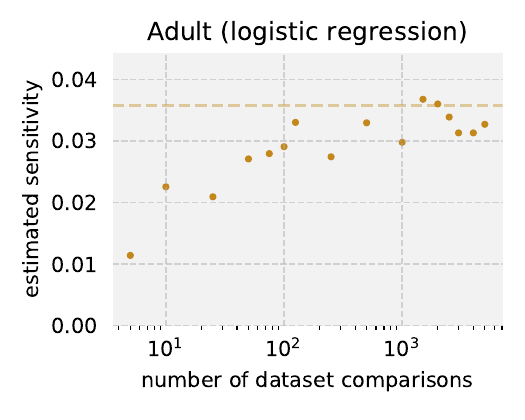}
    \end{minipage}
    \begin{minipage}{1\textwidth}
        \centering
        \includegraphics[scale=0.55]{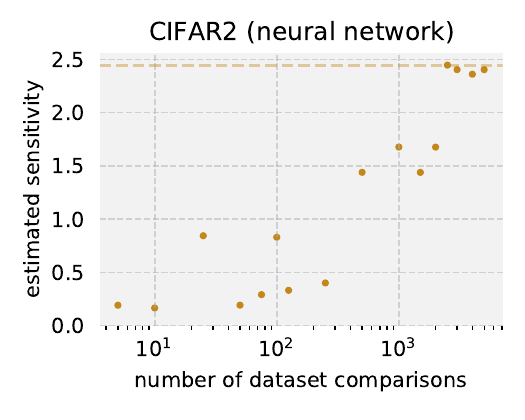}
        \includegraphics[scale=0.55]{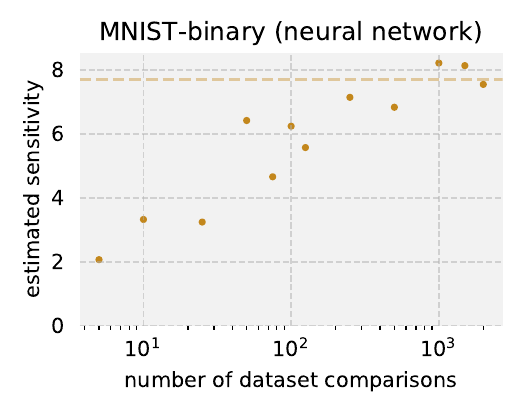}
        \includegraphics[scale=0.55]{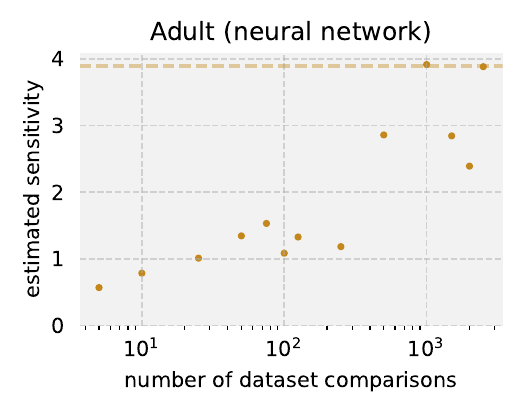}
    \end{minipage}
    \caption{We show how increasing the number of experiments used impact the estimates of empirical sensitivity $\empbound$ for all datasets and models.
    Dashed horizontal lines show the values of $\empbound$ used in the paper.
    Note that the x-axis is in log scale.}
    \label{fig:consistency_empbound}
\end{figure*}

\begin{figure*}[h]
        \centering
        \includegraphics[scale=0.55]{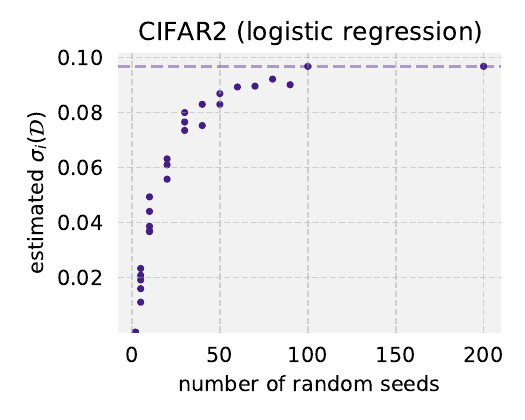}
        \includegraphics[scale=0.55]{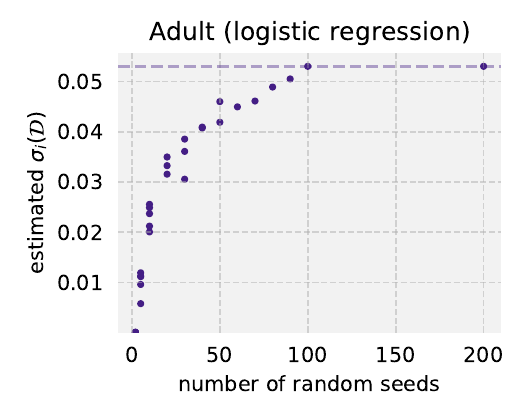}
    \caption{We show how increasing the number of seeds used impact the estimates of $\sigmai$ for all datasets, for logistic regression (we don't estimate $\sigmai$ for neural networks).
    Dashed horizontal lines show the values of $\sigmai$ used in the paper.}
    \label{fig:consistency_sigma}
\end{figure*}

In Figures~\ref{fig:consistency_empbound} and \ref{fig:consistency_sigma} we report the stability of results as we increase the number of experiments, to complement Figures~\ref{fig:consistency} for the rest of the datasets.

\end{document}